\documentclass[12pt]{article}
\usepackage{url}
\usepackage{authblk}
\usepackage[utf8]{inputenc}
\usepackage[T1]{fontenc}
\usepackage[a4paper,margin=1in]{geometry}
\usepackage{setspace}
\usepackage{graphicx}
\usepackage{xcolor}
\usepackage{booktabs}
\usepackage{amssymb,amsmath}
\usepackage{orcidlink}

\usepackage{url}
\usepackage{hyperref}
\usepackage{tabularx}
\usepackage{newunicodechar}
\newunicodechar{│}{|}
\usepackage{newunicodechar}
\newunicodechar{├}{|}
\usepackage{newunicodechar}
\newunicodechar{─}{-}
\usepackage{subcaption}
\usepackage{float}
\usepackage{enumitem}
\usepackage{authblk}
\usepackage{caption}
\usepackage{tabularx} 

\hypersetup{
  colorlinks=true,
  linkcolor=blue,
  citecolor=blue,
  urlcolor=blue,
  pdfauthor={- Authors},
  pdftitle={DVS-PedX: Synthetic-and-Real Event-Based Pedestrian Data for Detection and Crossing Intention}
}

\newcommand{\orcidauthorA}{0000-0003-3941-6065}
\newcommand{\orcidauthorE}{0000-0002-4401-2957}

\title{\textbf{DVS-PedX: Synthetic-and-Real Event-Based Pedestrian Dataset}}

\author[1]{Mustafa Sakhai\orcidlink{\orcidauthorA}}
\author[1]{Kaung Sithu}
\author[1]{Min Khant Soe Oke}
\author[1,2]{Maciej Wielgosz\orcidlink{\orcidauthorE}}

\affil[1]{Faculty of Computer Science, Electronics and Telecommunications, 
AGH University of Science and Technology, 30-059 Krakow, Poland\\
\texttt{msakhai@agh.edu.pl} (M.S.);\;
\texttt{sithu@student.agh.edu.pl} (K.S.)\\
\texttt{oke@student.agh.edu.pl} (M.K.S.O.);\;
\texttt{wielgosz@agh.edu.pl} (M.W.)}

\affil[2]{Academic Computer Centre AGH, AGH University of Science and Technology,
30-950 Krakow, Poland}

\affil[*]{Correspondence: \texttt{msakhai@agh.edu.pl} (M.S.)}

\date{} 

\begin{document}
\maketitle

\begin{abstract}
Event cameras like Dynamic Vision Sensors (DVS) report micro-timed brightness changes instead of full frames, offering low latency, high dynamic range, and motion robustness. \textbf{DVS-PedX} (Dynamic Vision Sensor Pedestrian eXploration) is a neuromorphic dataset designed for pedestrian detection and crossing-intention analysis in normal and adverse weather conditions across two complementary sources: (1) synthetic event streams generated in the CARLA simulator for controlled ``approach--cross'' scenes under varied weather and lighting; and (2) real-world JAAD dash-cam videos converted to event streams using the \textit{v2e} tool, preserving natural behaviors and backgrounds. Each sequence includes paired RGB frames, per-frame DVS ``event frames'' (33\,ms accumulations), and frame-level labels (crossing vs.\ not crossing). We also provide raw AEDAT~2.0/AEDAT~4.0 event files and AVI DVS video files and metadata for flexible re-processing. Baseline spiking neural networks (SNNs) using SpikingJelly illustrate dataset usability and reveal a sim-to-real gap, motivating domain adaptation and multimodal fusion. DVS-PedX aims to accelerate research in event-based pedestrian safety, intention prediction, and neuromorphic perception.
\end{abstract}

\section*{Background \& Summary}
Dynamic Vision Sensors (DVS), also known as event cameras, asynchronously capture brightness changes rather than outputting full image frames at a fixed rate, producing streams of events with microsecond temporal resolution and high dynamic range. Each event encodes pixel location, time, and polarity of an intensity change. These properties help detect fast motion with minimal latency and reduced data redundancy, which is attractive for autonomous driving and intelligent transportation \cite{gallego2022survey}.

The DVS-PedX dataset provides event-based data for pedestrian detection and crossing intention analysis from both simulation and real video domains. It builds upon the Joint Attention in Autonomous Driving (JAAD) dataset \cite{rasouli2017}, a widely used collection of dashcam videos with rich annotations of pedestrian behaviors and crossing events of 346 short video clips (5--10\,s long) extracted from over 240 hours of driving footage, and augments it with event representations using the \textit{v2e} converter \cite{hu2021v2e}. We converted the full JAAD dataset from RGB to DVS and validated it using spiking neural networks (SNNs) presented in~\cite{sakhai2024electronics}. Complementing this, DVS-PedX includes a synthetic subset of event data generated using the CARLA driving simulator \cite{dosovitskiy2017carla}. By combining these two sources, DVS-PedX enables direct comparisons between synthetic and real (converted) event data for evaluating learning models and understanding sim-to-real transfer.

Event-based models are naturally aligned with spiking neural networks (SNNs). Prior work using portions of this dataset demonstrated SNNs for real-time crossing detection under adverse weather in simulation \cite{sakhai2024electronics}. These studies underscore the importance of providing both types of data, as done in DVS-PedX, to support research on robust neuromorphic perception and domain adaptation for pedestrian crossing scenarios during normal and adverse weather conditions.

\section*{Methods}
\subsection*{Synthetic Data Generation in Simulation}
 We prepared (11\,GB) the first component of DVS-PedX in a controlled environment using the CARLA simulator \cite{dosovitskiy2017carla}. A single autonomous vehicle approaches a pedestrian crosswalk; a pedestrian may or may not cross during each run. Conditions were randomized according to lighting (day, dusk, night) and weather (clear, rain, fog). The appearance of the pedestrians and the start time of the crossing were also randomized. Labels (\texttt{1} crossing / \texttt{0} not crossing) on the CARLA frames: A frame is positive if any pedestrian is actively crossing in that frame; otherwise, negative.

\begin{figure}[H]
  \centering
  \includegraphics[width=0.92\linewidth]{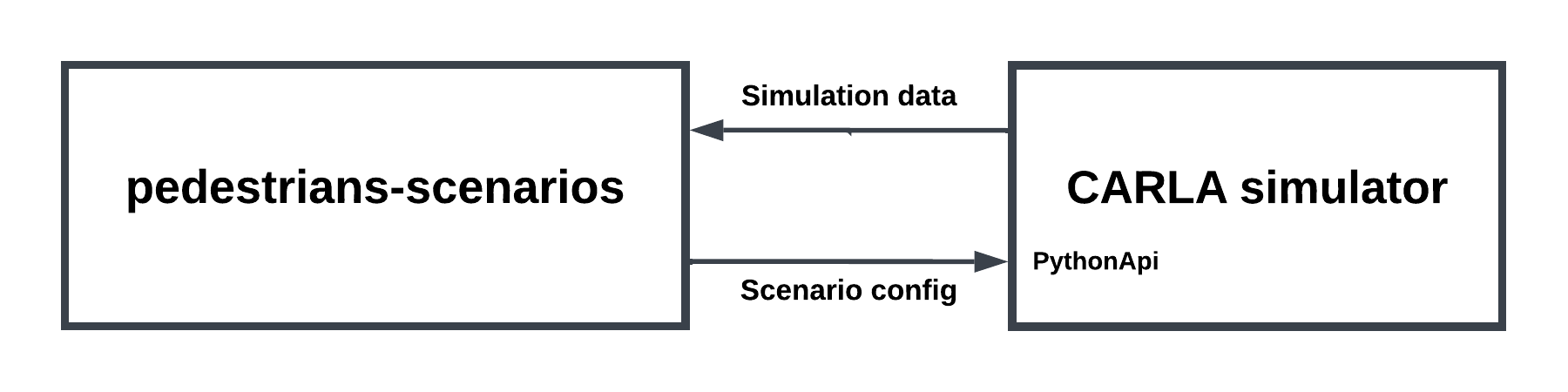}
  \caption{Data-generation loop used for the simulated sequences. Scenario configurations produced by \texttt{pedestrians-scenarios} are sent to the CARLA simulator via the PythonAPI (rightward arrow), which returns simulation data (leftward arrow) comprising sensor streams and metadata.}
  \label{fig:sim-data-loop}
\end{figure}

 As shown in Figure~\ref{fig:sim-data-loop}, the simulation workflow begins with \texttt{pedestrians-scenarios} generating scenario configurations, which are applied in CARLA through the PythonAPI. The simulator then returns the corresponding sensor streams and metadata, enabling reproducible variation in lighting and weather while keeping the crossing setup controlled.

Two virtual sensors were mounted from the driver's point of view: a standard RGB camera and a DVS event camera. The RGB camera recorded frames at 30\,Hz; the DVS produced an asynchronous stream of $(x,y,t,\text{polarity})$ events. We export a user-friendly ``event frame'' sequence by accumulating events within each 33\,ms interval, which produces a DVS frame aligned to the corresponding RGB frame (30\,Hz). These event frames encode motion, the moving edges of pedestrians and vehicles, while static backgrounds vanish.

\begin{figure}[H]
    \centering
\includegraphics[scale = 0.12 ]{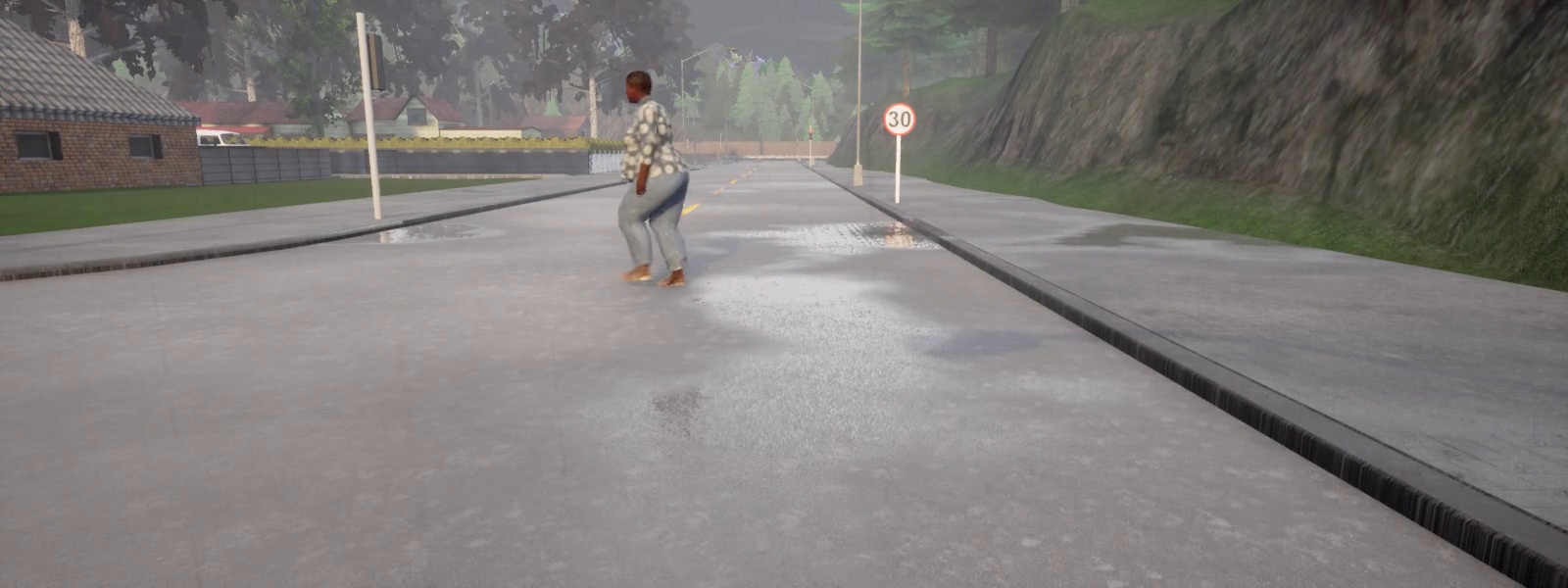}
\includegraphics[scale = 0.12 ]{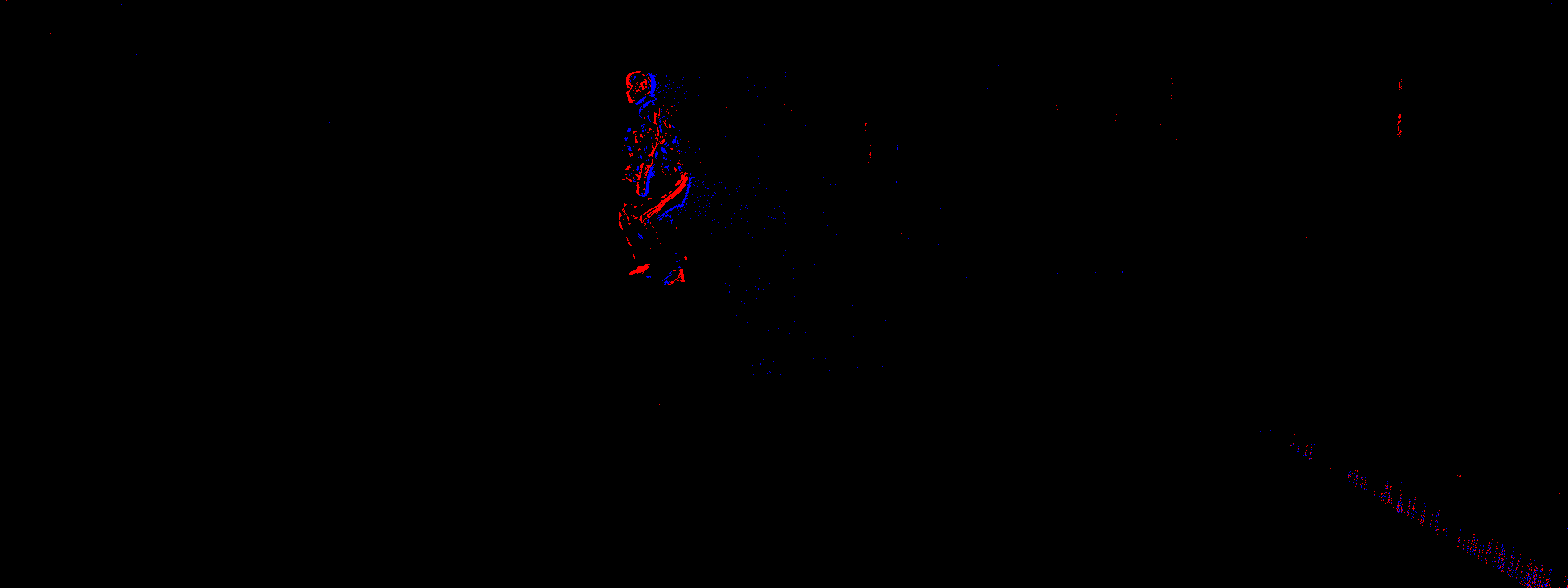}

\includegraphics[scale = 0.12 ]{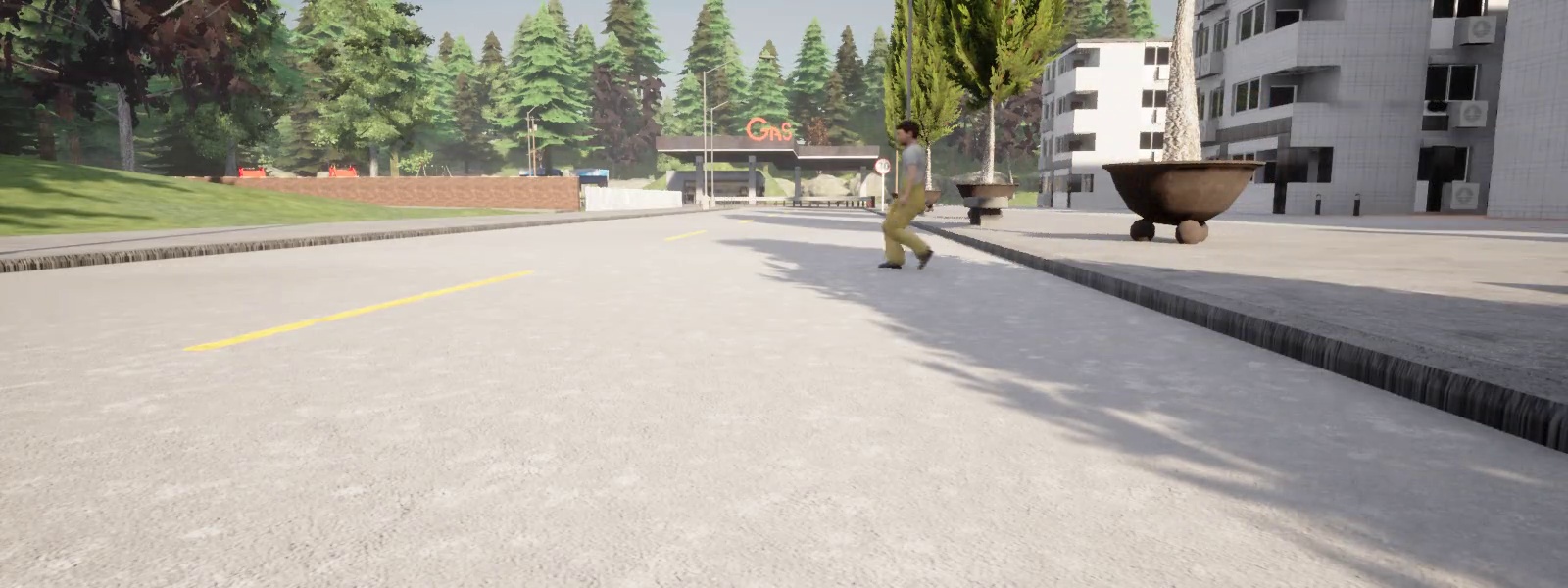}
\includegraphics[scale = 0.12 ]{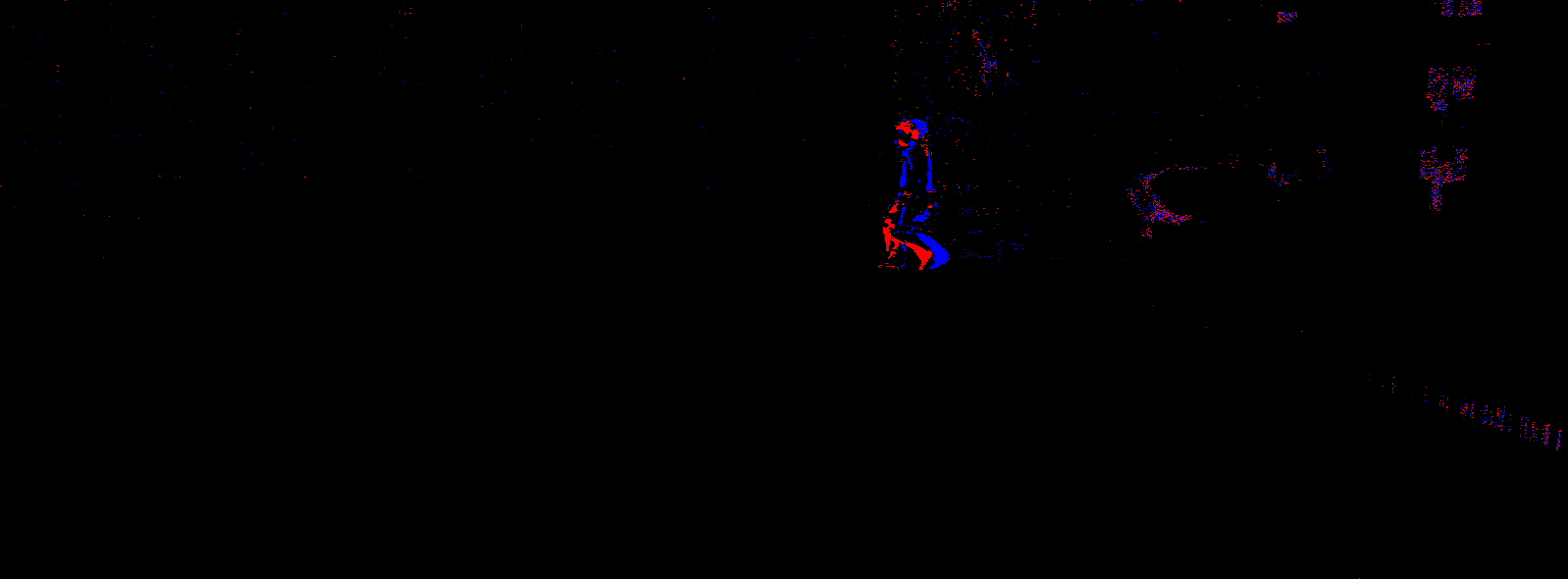}

\includegraphics[scale = 0.12 ]{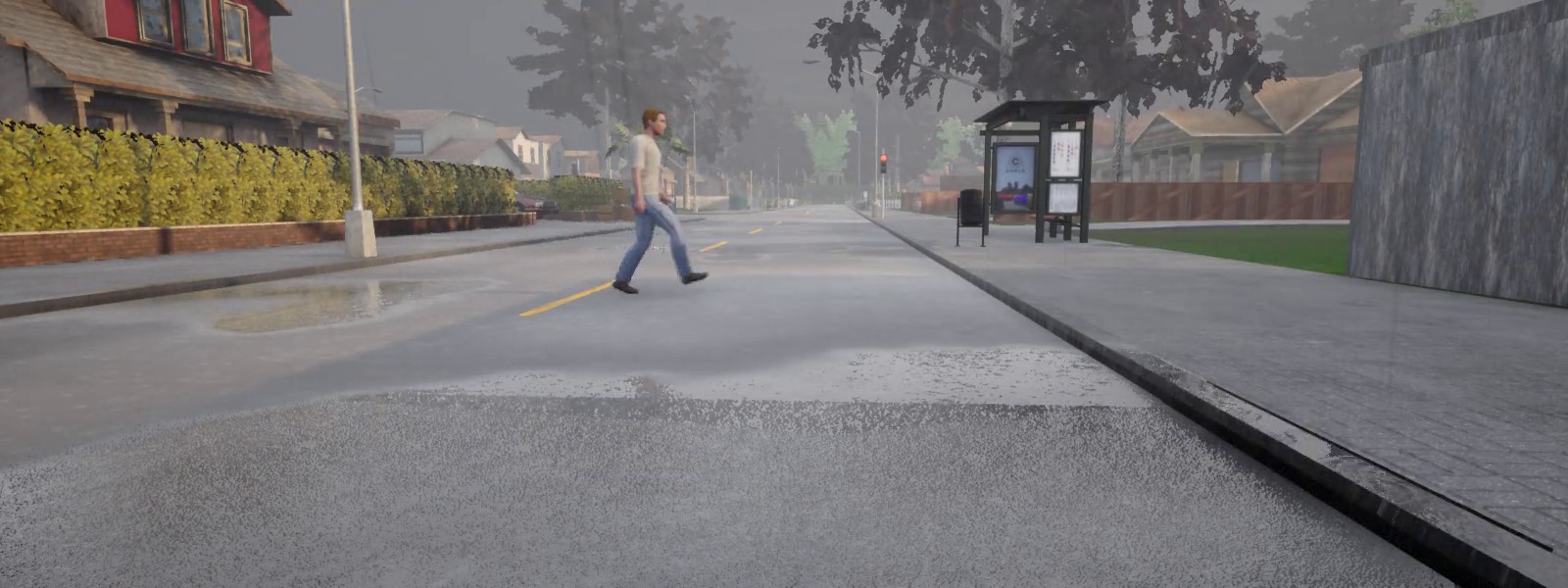}
\includegraphics[scale = 0.12 ]{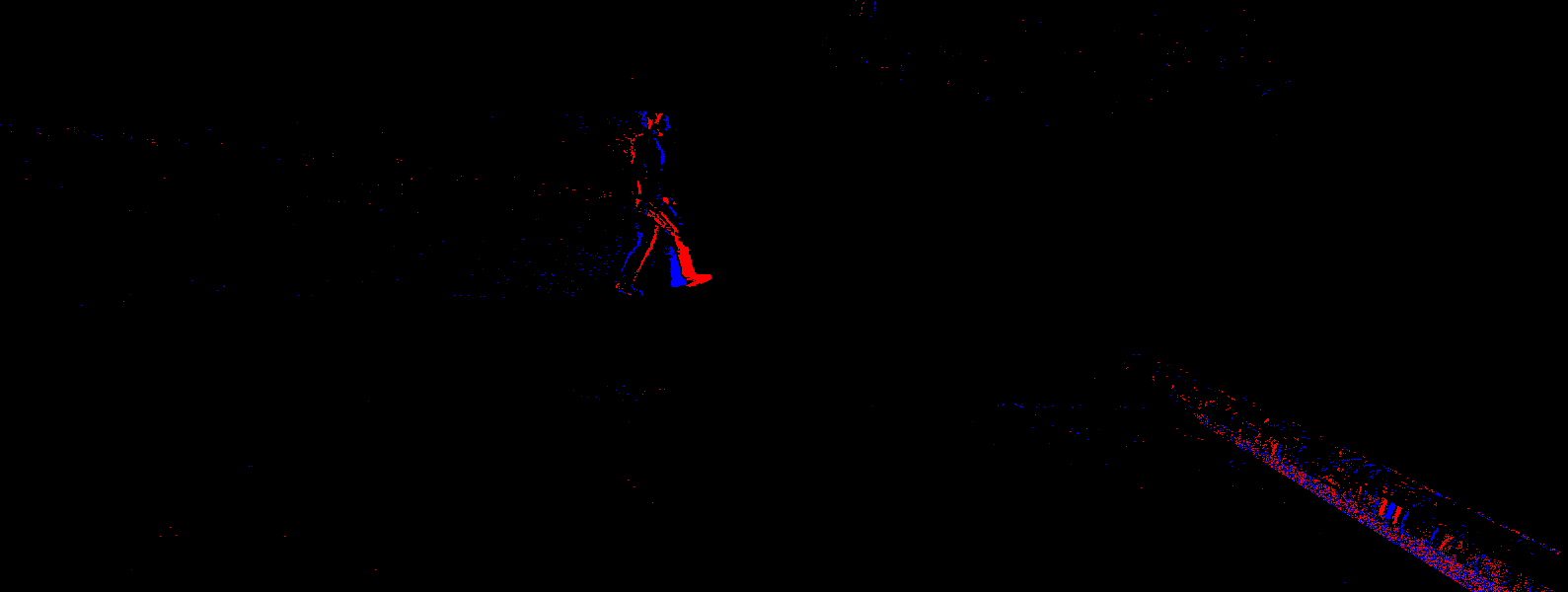}
    
    \caption{Examples of paired CARLA simulation samples showing the same pedestrian scene in RGB (left) and the corresponding DVS event-based representation (right). These comparisons highlight how temporal dynamics and motion edges are captured more distinctly in DVS streams.}
    \label{fig:carla-rgb-vs-dvs}
\end{figure}

As shown in Fig.~\ref{fig:carla-rgb-vs-dvs}, paired samples from CARLA demonstrate the correspondence between RGB appearance and DVS event representations, where motion-induced edges of the crossing pedestrian dominate the DVS frames. These examples illustrate the complementary cues captured by the two modalities across weather and lighting conditions.

\subsection*{Real Video Conversion to Event Data}
The second component is derived from JAAD \cite{rasouli2017}. JAAD contains 5--10\,s real-world dashcam clips at 30\,Hz, with rich annotations of pedestrian behaviors and whether crossing occurs. We converted all 346 JAAD clips to DVS (27GB) using \textit{v2e} \cite{hu2021v2e}, which emulates event generation from video by modeling pixel-level brightness change and, when beneficial, intermediate frame interpolation. Parameters were set analogously to the synthetic DVS setup (contrast threshold $\approx 0.3$, ON/OFF polarity). As with the simulated data, we provide per-frame DVS images at 30\,Hz by accumulating events between frames. 

\section*{Data Records}
\noindent \textbf{Public dataset link (Zenodo DOI): \href{https://doi.org/10.5281/zenodo.17030898}{10.5281/zenodo.17030898}}

\medskip
\noindent DVS-PedX is organized into two top-level folders: \texttt{carla\_simulator} and \texttt{JAAD\_DVS}. Within \texttt{carla\_simulator}, sequences are grouped by weather: \texttt{GoodWeather} (117 sequences) and \texttt{BadWeather} (81 sequences), and each sequence folder contains:

\renewcommand{\arraystretch}{1.2} 

\begin{table}[h!]
\centering
\begin{tabularx}{\linewidth}{lX}
\toprule
\textbf{Folder} & \textbf{Contents and Labeling Scheme} \\
\midrule
\texttt{frames\_dataset\_adverse\_weather\_rgb/} & 
RGB frames in JPEG format (e.g., \texttt{name-label.jpg}) captured under adverse weather. 
Labels: 0 = no pedestrian crossing, 1 = pedestrian crossing. \\

\texttt{frames\_dataset\_rgb/} & 
RGB frames in JPEG format captured under normal weather. 
Labeling scheme as above. \\

\texttt{frames\_dataset\_adverse\_weather\_dvs/} & 
DVS event frames in PNG format (e.g., \texttt{name-label.png}) accumulated under adverse weather. 
Labels: 0 = no pedestrian crossing, 1 = pedestrian crossing. \\

\texttt{frames\_dataset\_dvs/} & 
DVS event frames in PNG format accumulated under normal weather. 
Labeling scheme as above. \\
\bottomrule
\end{tabularx}
\caption{Organization of CARLA simulation data into RGB and DVS frame subsets under normal and adverse weather conditions.}
\label{tab:carla-seq-contents}
\end{table}

\noindent Within \texttt{JAAD\_DVS}, each folder (\texttt{video\_0001} to \texttt{video\_0346}) corresponds to a JAAD clip converted into DVS format. The files contained in each folder are summarized in Table~\ref{tab:jaad-dvs}.

\begin{table}[H]
\centering
\begin{tabular}{p{0.32\linewidth} p{0.62\linewidth}}
\toprule
\textbf{File} & \textbf{Contents and Description} \\
\midrule
\texttt{dvs-video-frame\_times.txt} & Text file containing per-frame timestamps (s) for the generated DVS video. \\

\texttt{dvs-video.avi} & AVI file of the converted DVS video derived from the original JAAD RGB clip. \\

\texttt{v2e-args.txt} & Command-line parameters used by the \textit{v2e} conversion tool. \\

\texttt{video\_NAME\_v2.aedat} & Raw events in AEDAT~2.0 format. \\

\texttt{video\_NAME\_v4.aedat4} & Raw events in AEDAT~4.0 format. \\
\bottomrule
\end{tabular}
\caption{Contents of each JAAD-DVS converted folder. Each folder corresponds to one JAAD clip and contains converted video, timestamps, raw event files (AEDAT~2.0/4.0), and conversion parameters.}
\label{tab:jaad-dvs}
\end{table}

\begin{figure}[H]
  \centering
  \includegraphics[width=1.2\linewidth]{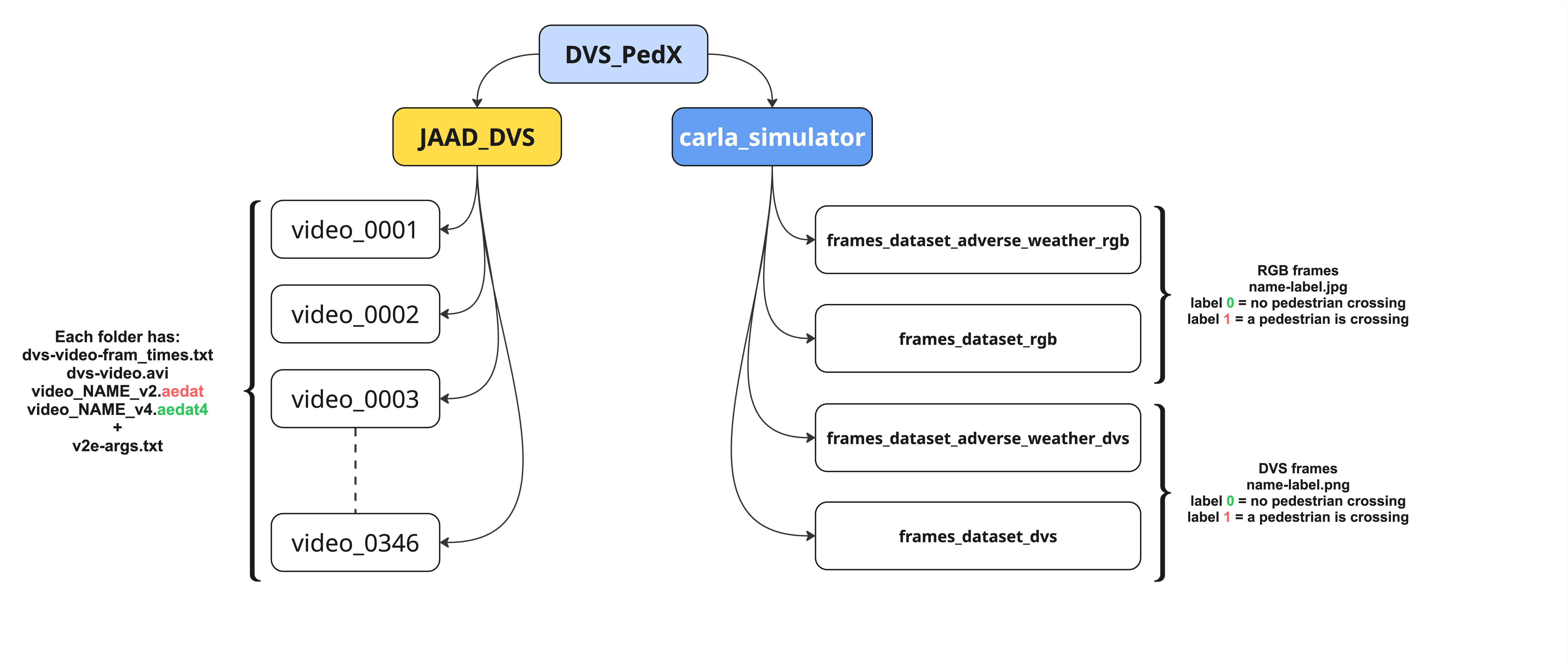}
  \caption{Directory structure of the DVS-PedX dataset. JAAD\_DVS clips each contain v2e outputs (\texttt{dvs-video.avi}, AEDAT~2.0/4.0, etc.), while CARLA sequences are split into RGB and DVS frames.}
  \label{fig:dataset-arch}
\end{figure}

The overall folder hierarchy and file organization of DVS-PedX are summarized in Fig.~\ref{fig:dataset-arch}.

\noindent Event PNGs map positive events to brighter intensities and negative events to darker intensities about a mid-level gray. For precise timing analysis or alternative accumulation windows, use the provided AEDAT files. 


\begin{figure}[H]
  \centering
  \begin{subfigure}[b]{0.45\linewidth}
    \includegraphics[width=\linewidth]{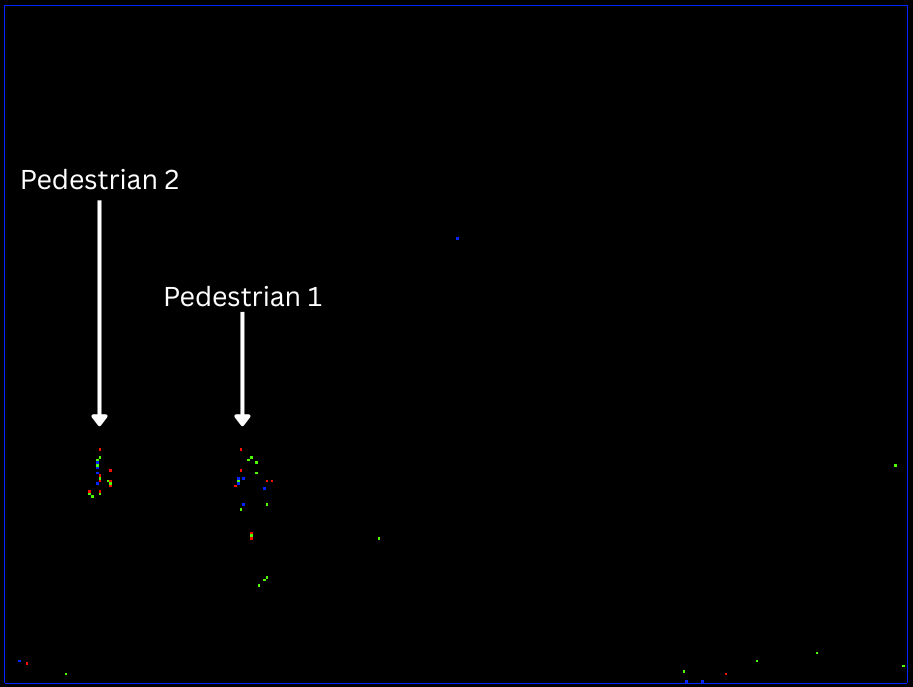}
    \caption{Pedestrian sequence (1)}
  \end{subfigure}
  \hfill
  \begin{subfigure}[b]{0.45\linewidth}
    \includegraphics[width=\linewidth]{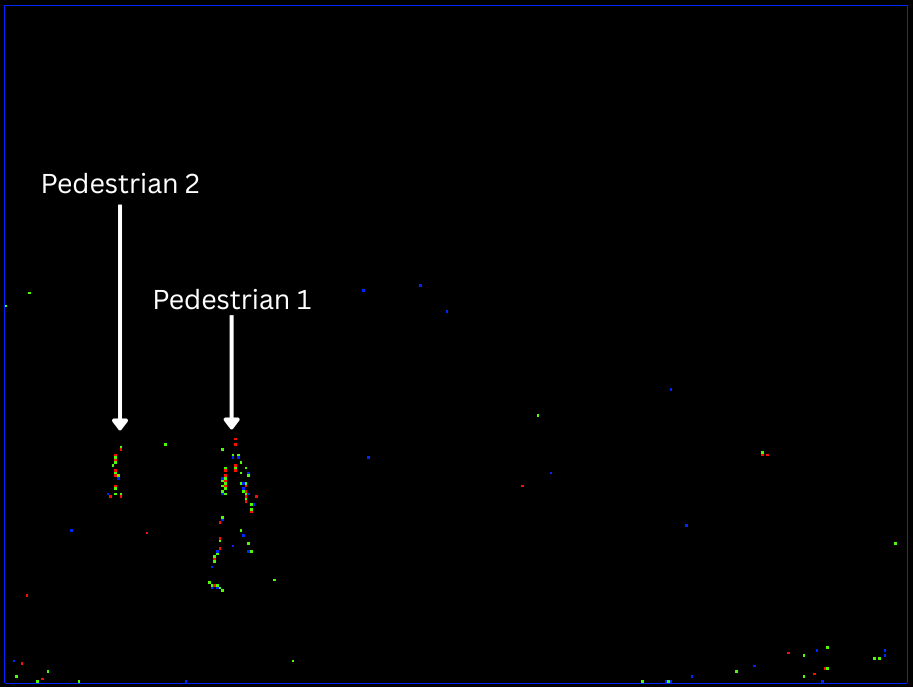}
    \caption{Pedestrian sequence (2)}
  \end{subfigure}

  \vskip\baselineskip

  \begin{subfigure}[b]{0.45\linewidth}
    \includegraphics[width=\linewidth]{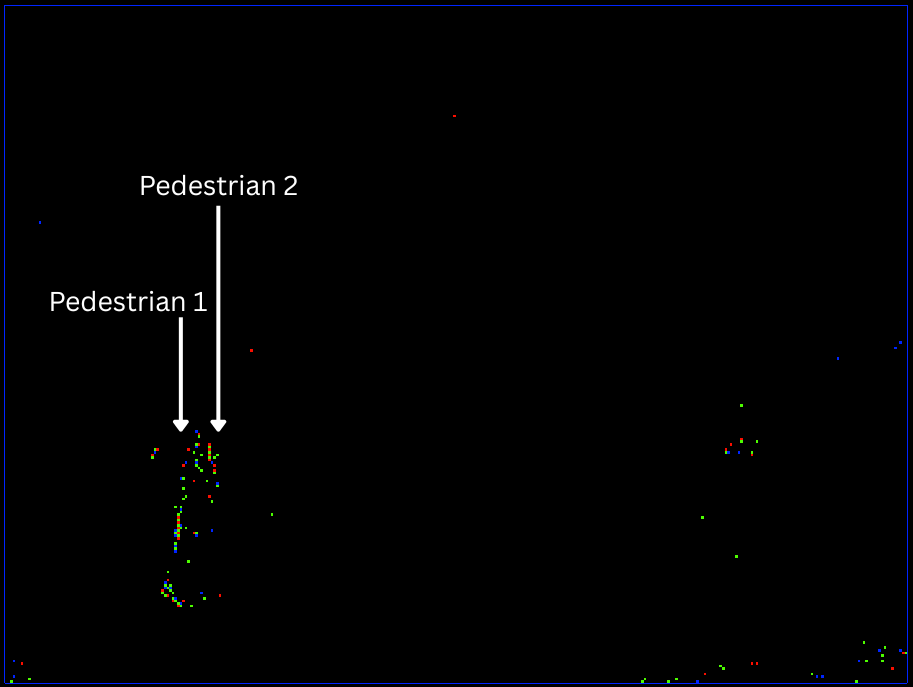}
    \caption{Pedestrian sequence (3)}
  \end{subfigure}
  \hfill
  \begin{subfigure}[b]{0.45\linewidth}
    \includegraphics[width=\linewidth]{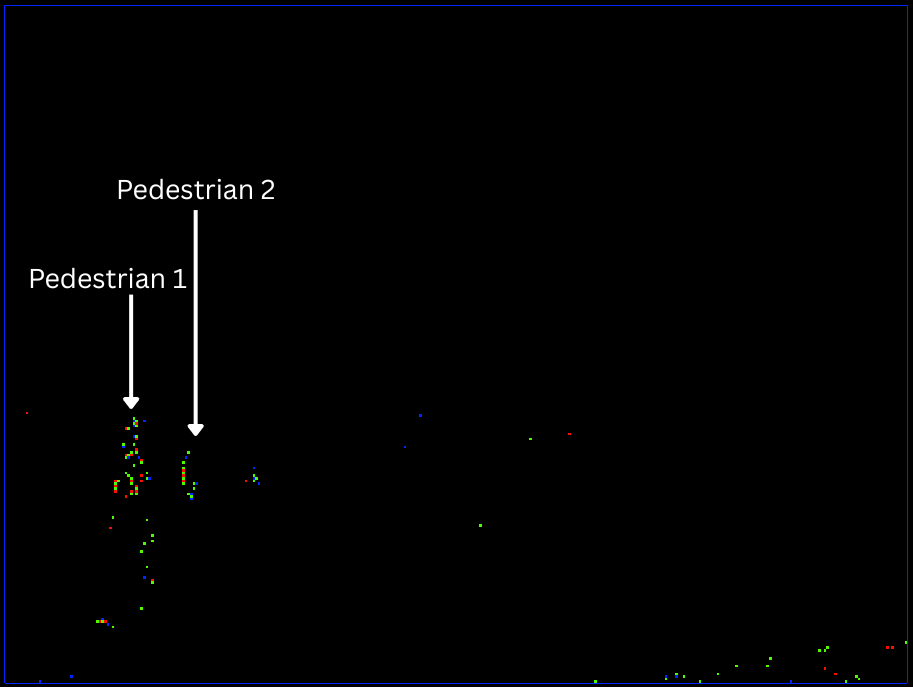}
    \caption{Pedestrian sequence (4)}
  \end{subfigure}

  \caption{Sequence of JAAD--DVS frames from the DVS-PedX dataset, showing two pedestrians moving across the scene. Subfigures (a–d) illustrate consecutive frames in the event stream.}
  \label{fig:dvs-pedestrians-sequence}
\end{figure}

As illustrated in Fig.~\ref{fig:dvs-pedestrians-sequence}, a sequence of four DVS frames shows two pedestrians moving across the scene.

\section*{Data Overview}
\begin{itemize}[leftmargin=2em]
  \item \textbf{Synthetic (CARLA):} 198 sequences (117 ``good weather'', 81 ``bad weather''), each 30\,s @ 30\,Hz $\Rightarrow$ 900 frames per sequence per modality.
  \item \textbf{Real-converted (JAAD→DVS via v2e; JAAD-DVS):} 346 video sequences (5--10\,s each).
  \item \textbf{Labels:} Frame-level binary crossing; positives are a minority (realistic class imbalance).
  \item \textbf{Modalities:} Aligned RGB and DVS event frames; raw AEDAT provided for precise timing.
\end{itemize}

\section*{Technical Validation}
For simulation, DVS parameters were chosen to emulate typical event camera behavior (distinct ON/OFF polarities, absence of motion blur). We verified event activity patterns: near-zero events when actors are static and spikes during motion, consistent with DVS properties \cite{gallego2022survey}. For JAAD-DVS, visual inspection confirmed events align with obvious motions. It is suitable for algorithmic evaluation; users may inject noise if needed \cite{hu2021v2e}.

\textbf{Label consistency.} Simulation labels follow controlled ground truth (crossing interval precisely known). JAAD-derived labels rely on published annotations; ambiguous ``approach but no cross'' cases remain negative for consistency with outcomes \cite{rasouli2017}.

\textbf{Baseline generalization.} A three-layer convolutional SNN using SpikingJelly \cite{zhou2023spikingjelly} trained on the synthetic subset achieves high synthetic-test accuracy; performance drops on JAAD-DVS without adaptation and improves after mixing real-converted data---consistent with sim-to-real gaps and prior reports \cite{sakhai2024electronics}. This validates that DVS-PedX is realistic and non-trivial.

\begin{figure}[H]
  \centering
  \includegraphics[width=\linewidth]{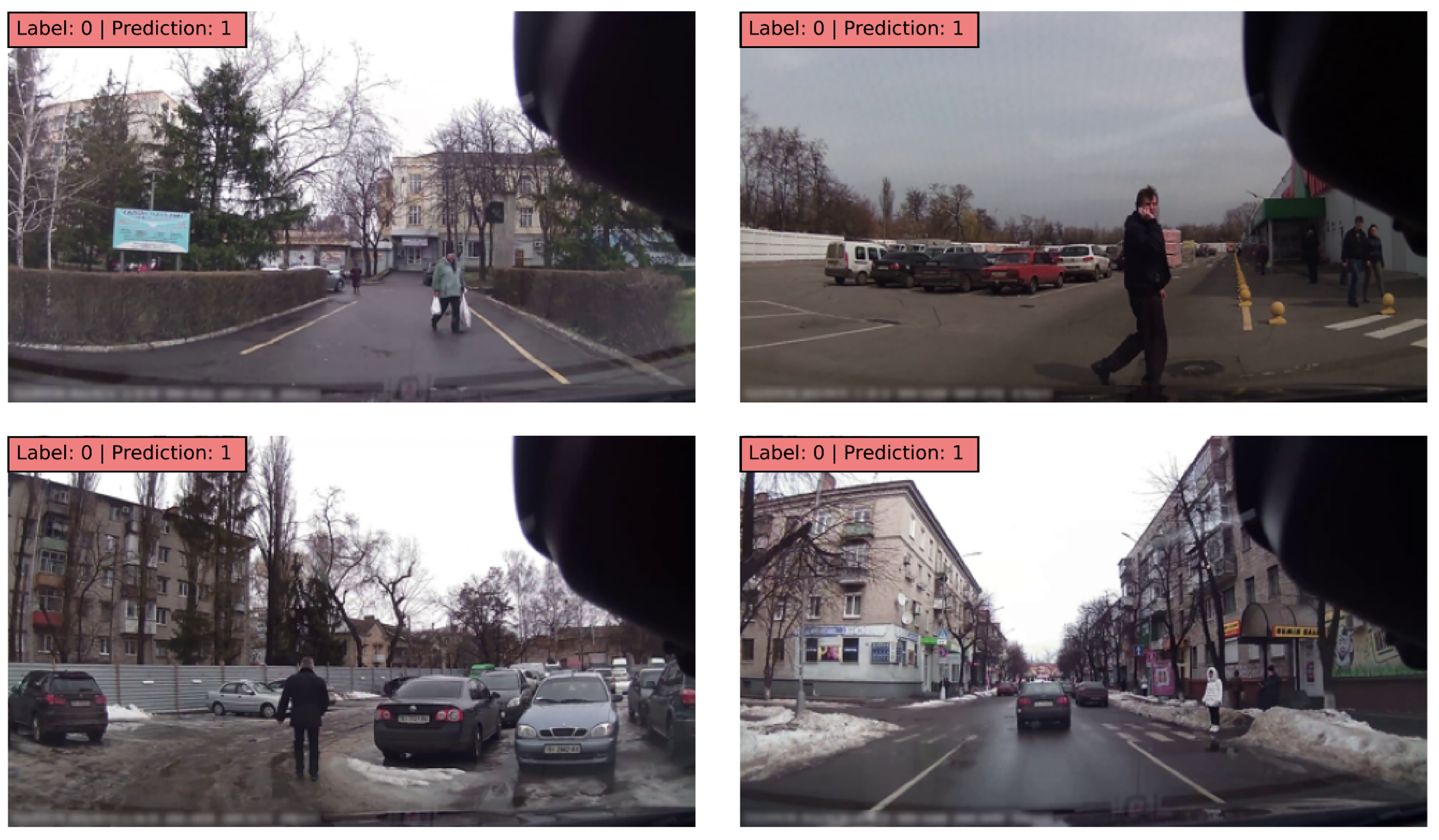}
  \caption{Examples of incorrectly classified frames evaluated on the bad-weather subset of the JAAD dataset. Predictions were generated using the SPS R18T model in a single-frame classification task. In these cases, the frame-level labels did not fully capture the pedestrian’s behavior, leading to apparent misclassifications~\cite{sakhai2024electronics}.}
  \label{fig:jaad-classification}
\end{figure}

As shown in Fig.~\ref{fig:jaad-classification}, some apparent misclassifications arise because frame-level labels do not always align with pedestrian intent, underscoring the importance of sequence-level modeling and temporal context in this dataset.

\begin{figure}[H]
  \centering
  \includegraphics[width=\linewidth]{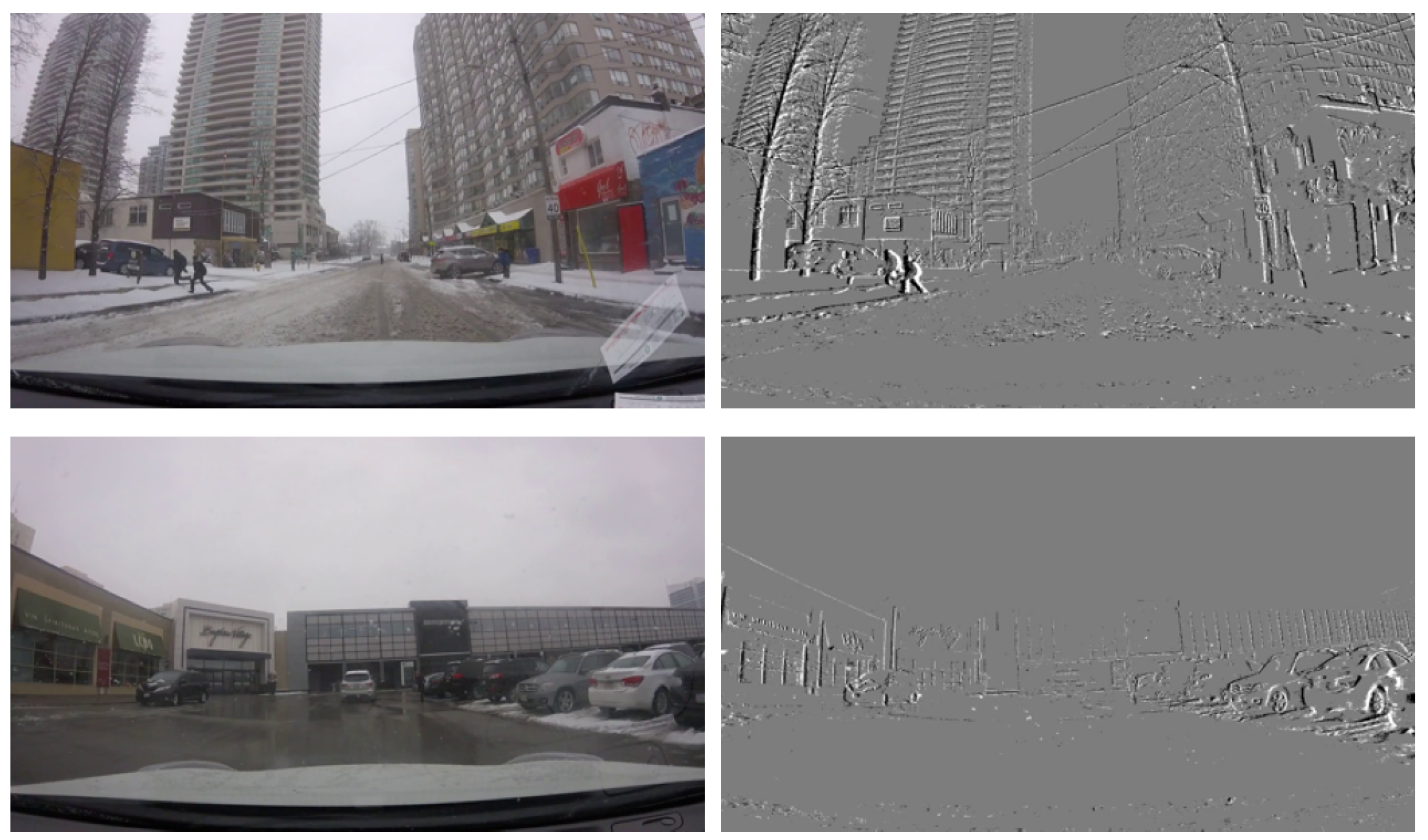}
  \caption{Examples of RGB-to-DVS data conversion of frames from the JAAD dataset. The converted DVS representations highlight motion-induced edges and structural contours, showing the potential of such methods to impute missing DVS modalities in datasets that contain only RGB recordings.}
  \label{fig:jaad-rgb-dvs}
\end{figure}

\medskip
To evaluate synthetic DVS data, JAAD RGB videos~\cite{rasouli2017} were converted to event-based formats using the v2e tool, and the performance of an SNN model was tested and published in~\cite{sakhai2024electronics}. The DMT22 dataset~\cite{DMT22-Zenodo}, containing real DVS events and synchronized APS intensity (grayscale) frames, enabled direct comparisons between synthetic DVS, real DVS, and frame-based intensity data, validating synthetic event-based vision as a practical alternative to physical DVS sensors. As shown in Fig.~\ref{fig:jaad-rgb-dvs}, RGB frames from JAAD can be converted into DVS-like representations that emphasize motion and edge information. This demonstrates the feasibility of augmenting RGB-only datasets with synthetic DVS modalities for neuromorphic research.

\begin{figure}[H]
  \centering
  \includegraphics[width=0.95\linewidth]{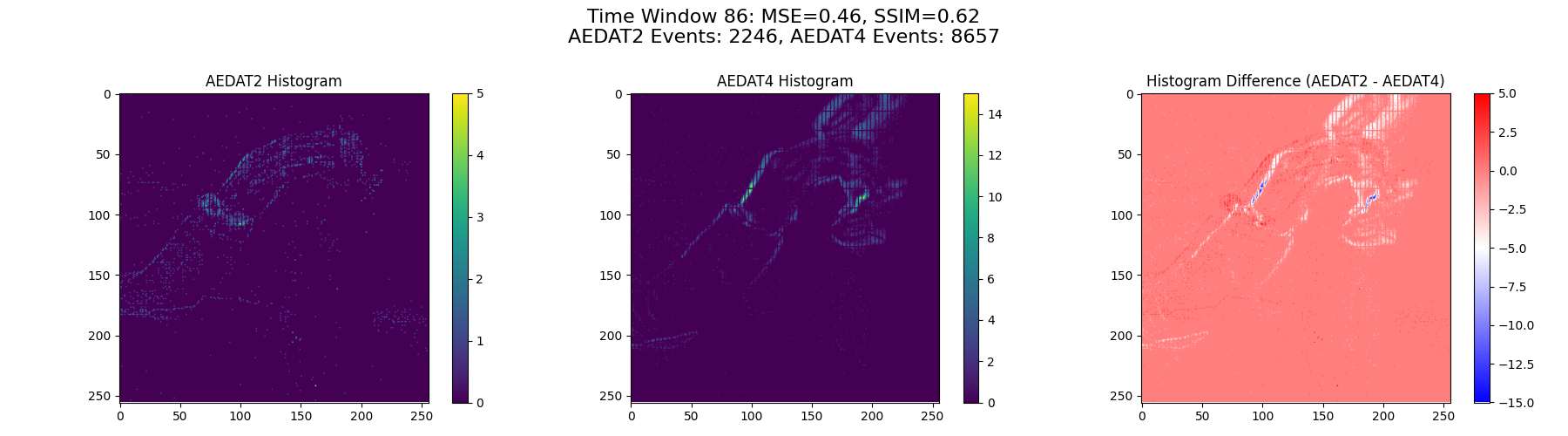}
  \caption{Visual comparison between original AEDAT 2.0 (Original DVS from DMT22) frames and synthetic AEDAT 4.0 (Converted DMT22 to DVS) frames, with histogram differences highlighting structural similarities and deviations.}
  \label{fig:aedat2-vs-aedat4}
\end{figure}

As illustrated in Fig.~\ref{fig:aedat2-vs-aedat4}, the comparison between original AEDAT 2.0 and synthetic AEDAT 4.0 frames highlights their structural similarities and differences. 

\begin{figure}[H]
  \centering
  \includegraphics[width=\linewidth]{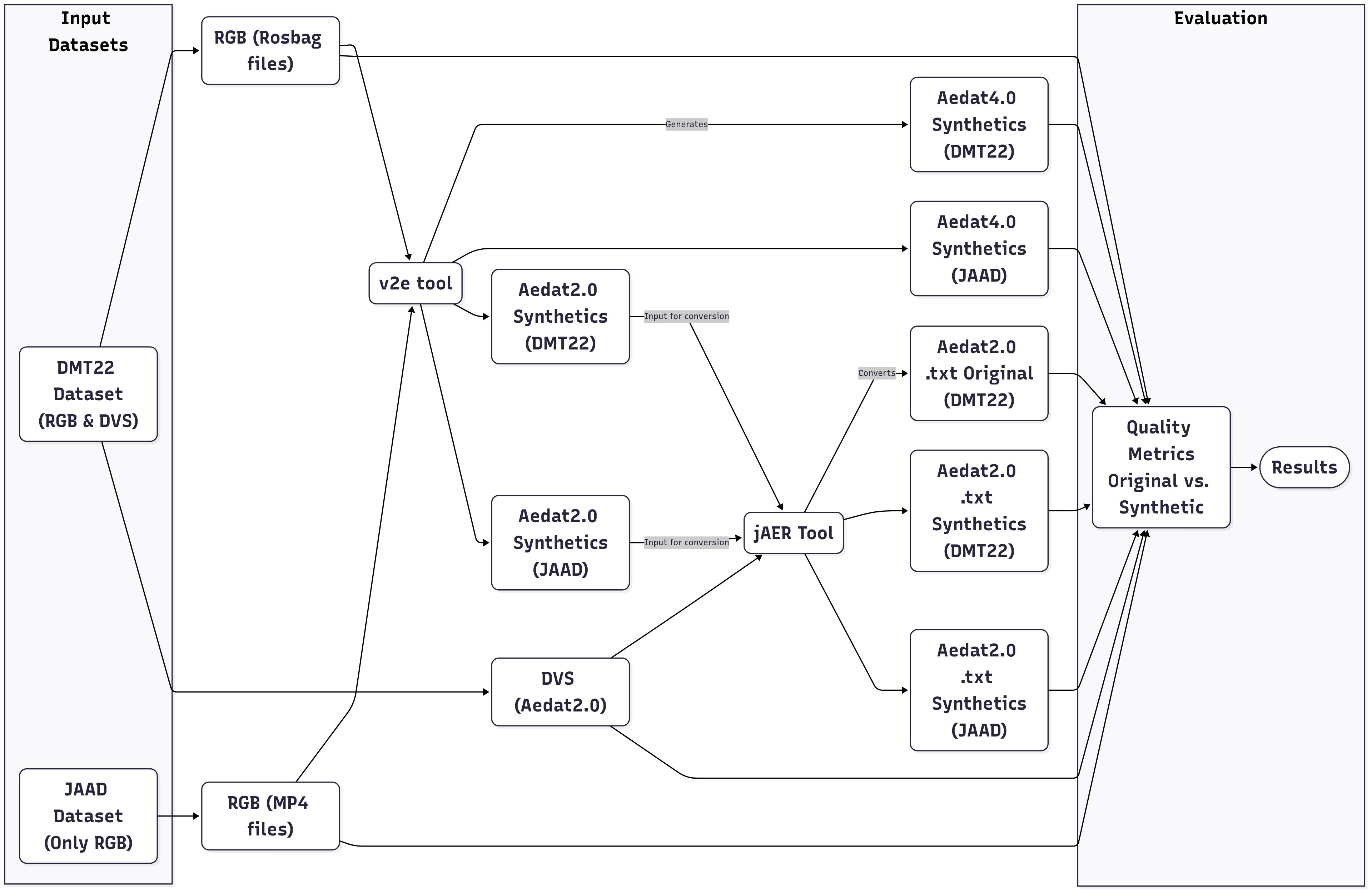}
  \caption{Frame-to-events conversion and validation pipeline used in the dataset. The diagram illustrates how RGB inputs from JAAD~\cite{rasouli2017} and APS intensity (grayscale) inputs from DMT22~\cite{DMT22-Zenodo} are transformed into synthetic DVS streams and processed for quality evaluation; original DMT22 event files are also exported to text via jAER for comparison.}
  \label{fig:rgb-to-dvs-pipeline}
\end{figure}

\noindent As summarized in Fig.~\ref{fig:rgb-to-dvs-pipeline}, frame-based streams—APS intensity (grayscale) for DMT22 and RGB for JAAD—are converted into synthetic event streams (AEDAT 2.0/4.0) using the same parameterized pipeline, while the original DMT22 event files are exported to text via jAER. All variants---original APS intensity frames, real DVS, and synthetic DVS---are then fed through a common quality-metrics module to enable like-for-like comparisons and aggregate results.

\section*{Usage Notes}

Table~\ref{tab:dataset-guidelines} summarizes the dataset usage guidelines, including recommended splits, event timing, and handling of class imbalance. These considerations provide practical directions for training, validation, and fusion strategies when working with RGB and DVS modalities.

\begin{table}[h!]
\centering
\begin{tabular}{p{0.22\linewidth} p{0.70\linewidth}}
\hline
\textbf{Tasks} & frame-level crossing detection, early intention prediction, domain adaptation (CARLA$\rightarrow$JAAD), modality comparisons (RGB vs. DVS vs. fusion). \\
\hline
\textbf{Recommended splits} & Train in CARLA (good + bad weather), validate in held CARLA, and \emph{test} in JAAD-DVS to measure domain transfer; optionally mix a small JAAD-DVS subset for fine-tuning. \\
\hline
\textbf{Event timing} & For precise temporal modeling, use AEDAT and build microslices (e.g., 1--5\,ms). \\
\hline
\textbf{Class imbalance} & Use loss weighting or sampling. \\
\hline
\textbf{Fusion} & Try dual-stream (RGB + DVS) architectures to combine appearance and motion signals. \\
\hline
\end{tabular}
\caption{Guidelines for splits, timing, class imbalance, and fusion.}
\label{tab:dataset-guidelines}
\end{table}

\section*{Data Availability}
\noindent \textbf{DVS-PedX:} \url{https://doi.org/10.5281/zenodo.17030898} \\
\noindent \textbf{Zenodo DOI: 10.5281/zenodo.17030898}

\noindent The dataset~\cite{sakhai2025dvspedx} is available with a persistent identifier and rich metadata (license: CC BY 4.0).

\section*{Code Availability}
\noindent \textbf{Code repository URL:} \url{https://github.com/MustafaSakhai/videos-specifications-rgb_to_dvs-v2e} \\
conversion (\textit{v2e} batch wrapper). Data license: CC BY 4.0. \textbf{Please cite this descriptor and the dataset DOI (10.5281/zenodo.17030898).}

\section*{Author Contributions}
\textbf{Conceptualization:} M.S.\quad
\textbf{Methodology:} M.S.\quad
\textbf{Software:} M.S.\quad
\textbf{Data curation:} M.S.\quad
\textbf{Validation:} K.S., M.K.S.O., M.S.\quad
\textbf{Visualization:} M.S.\quad
\textbf{Writing—original draft:} M.S.\quad
\textbf{Writing—review \& editing:} M.S., M.W.\quad
\textbf{Supervision:} M.W., M.S.

\section*{Competing Interests}
The authors declare no competing interests.

\section*{Acknowledgements}
We thank the developers of CARLA and \textit{v2e}, and the creators of JAAD for making their resources publicly available. We also thank \textbf{Krzysztof Jędrejasz} for his assistance in creating the CARLA dataset during his master’s studies, where part of this work formed his master’s thesis. Finally, we acknowledge computational support from ACC Cyfronet AGH (Prometheus cluster).


\end{document}